\documentclass[10pt,twocolumn,letterpaper]{article}

\usepackage{iccv}
\usepackage{times}
\usepackage{epsfig}
\usepackage{graphicx}
\usepackage{amsmath}
\usepackage{amssymb}

\usepackage{threeparttable}
\usepackage{multirow}
\usepackage{longtable}
\usepackage{rotating}
\usepackage{tabularx}
\usepackage{pifont}
\usepackage{bbding}
\usepackage{booktabs}
\usepackage{color}
\usepackage{enumitem}

\usepackage{bbm}
\usepackage{amssymb}
\usepackage{arydshln}
\usepackage{subfig}
\usepackage{bm}
\usepackage{axessibility}

\usepackage[pagebackref=true,breaklinks=true,letterpaper=true,colorlinks,bookmarks=false]{hyperref}

\iccvfinalcopy 

\def\iccvPaperID{6078} 

\ificcvfinal\fi

\begin{document}
	
	\title{Learn to Match: Automatic Matching Network Design for Visual Tracking}
	
	\author{\href{http://zhipengzhang.cn/}{\textcolor{black}{Zhipeng Zhang$^{1,2}$}}, ~Yihao Liu$^{2}$,~\href{https://sites.google.com/view/xiaowanghomepage/}{\textcolor{black}{Xiao Wang$^{3}$}}, ~Bing Li$^{1,2, \dagger}$, ~and Weiming Hu$^{1,2,4,\dagger}$\\
$^1$ National Laboratory of Pattern Recognition, Institute of Automation, Chinese Academy of Sciences \\
$^2$ School of AI, University of Chinese Academy of Sciences \quad $^3$ Peng Cheng Laboratory \\
$^4$ CAS Center for Excellence in Brain Science and Intelligence Technology
}

	
	\maketitle
	\ificcvfinal\thispagestyle{empty}\fi
	
	\begin{abstract}

    Siamese tracking has achieved groundbreaking performance in recent years, where the essence is the efficient matching operator cross-correlation and its variants. Besides the remarkable success, it is important to note that the heuristic matching network design relies heavily on expert experience. Moreover, we experimentally find that one sole matching operator is difficult to guarantee stable tracking in all challenging environments. Thus, in this work, we introduce six novel matching operators \textbf{from the perspective of feature fusion instead of explicit similarity learning}, namely Concatenation, Pointwise-Addition, Pairwise-Relation, FiLM, Simple-Transformer and Transductive-Guidance, to explore more feasibility on matching operator selection. The analyses reveal these operators' selective adaptability on different environment degradation types, which inspires us to combine them to explore complementary features. To this end, we propose binary channel manipulation (BCM) to search for the optimal combination of these operators. BCM determines to retrain or discard one operator by learning its contribution to other tracking steps. By inserting the learned matching networks to a strong baseline tracker Ocean \cite{Ocean}, our model achieves favorable gains by $67.2 \rightarrow 71.4$, $52.6 \rightarrow 58.3$, $70.3 \rightarrow 76.0$ success on OTB100, LaSOT, and TrackingNet, respectively. Notably, Our tracker, dubbed \textbf{AutoMatch}, uses less than half of training data/time than the baseline tracker, and runs at 50 FPS using PyTorch. Code and model are released at \href{https://github.com/JudasDie/SOTS}{https://github.com/JudasDie/SOTS}.

	\end{abstract}
	
\newcommand\blfootnote[1]{%
\begingroup 
\renewcommand\thefootnote{}\footnote{#1}%
\addtocounter{footnote}{-1}%
\endgroup 
}
{
	\blfootnote{Email: zhangzhipeng2017@ia.ac.cn \quad
 ~$^\dagger$ Corresponding authors.
	}
}

    
	
	\section{Introduction}

Generic object tracking, aiming to infer the location and scale of an arbitrary object in a video sequence, is one of the fundamental problems in computer vision \cite{intro1,intro2,SURVEY,intro3}. The recent prevailing Siamese methods \cite{siamFC, SiamBAN, SiamCAR, siamRPNpp, SiamFC++, SiamAtt, Ocean}, decompose the tracking problem into a \emph{relation learning} task and a \emph{state estimation} task. In the former case, the goal is to measure the similarity between exemplar and candidate (search) images. The second task,  which is normally comprised of foreground classification and scale regression \cite{ATOM, siamRPNpp, Ocean}, is followed to estimate the target state. 
	
	   
	  \begin{figure}
        \begin{center}
        	\includegraphics[width=1\linewidth]{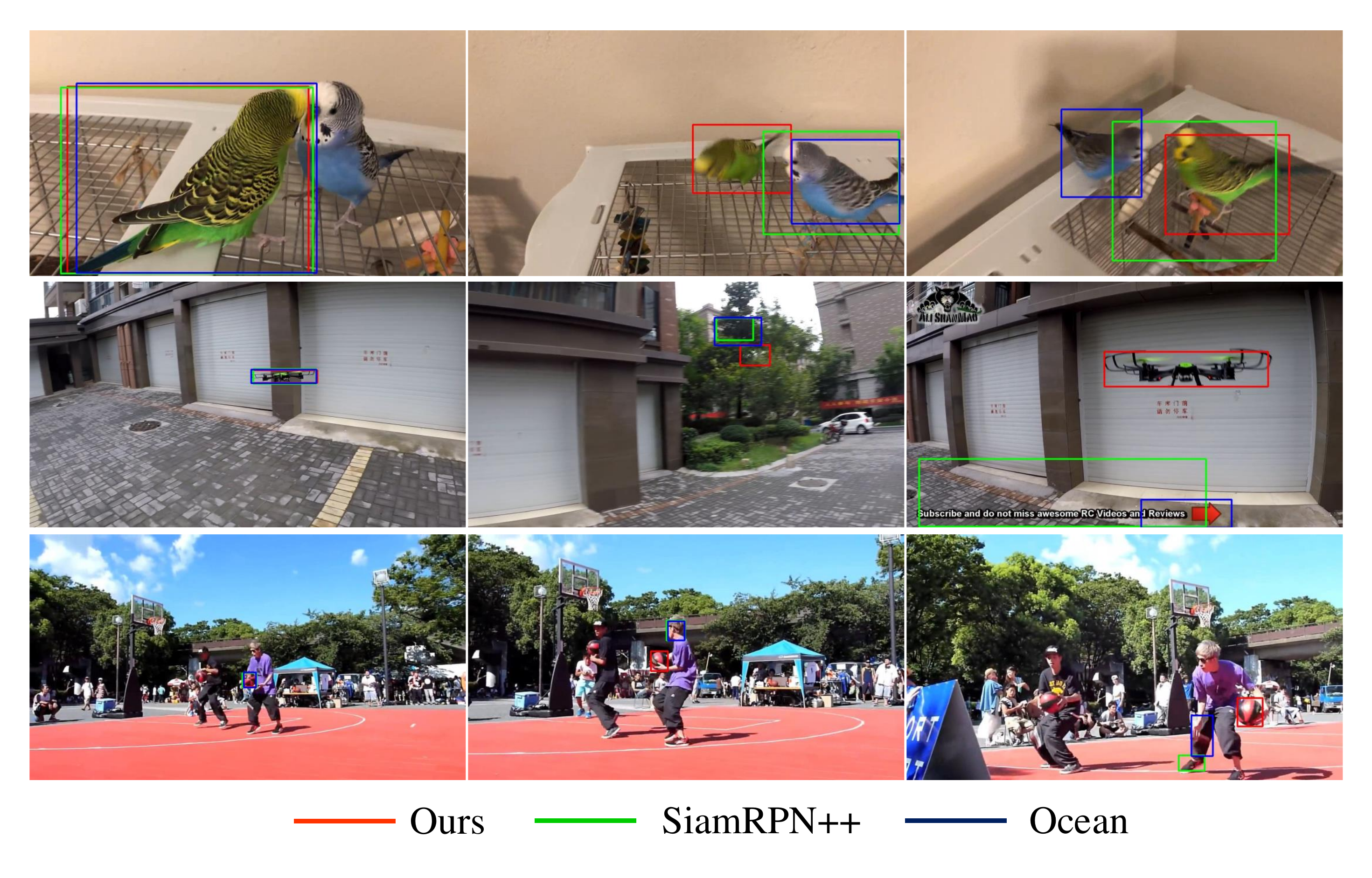}
        \end{center}
        \vspace{-1em}
        \caption{Comparisons of our approach with depthwise cross-correlation based trackers SiamRPN++ \cite{siamRPNpp} and Ocean \cite{Ocean}. Our model, employing the automatically searched matching networks, can better handle different challenging factors, \emph{e.g.,} distractor in the first video, occlusion and scale change of the second one, background clutter and fast motion of the third sequence.} 
    
        \label{fig:vis}
        \end{figure}

	 Fuelled by the emergence of object detection that facilitates bounding box regression, the network design for state estimation has substantially advanced in recent years \cite{SiamBAN, ATOM, siamRPN, SiamFC++, Ocean}. However, the advancements in relation learning have been limited. Previous works generally perform relation learning with heuristically designed matching operators. Concretely, the seminal work SiamFC \cite{siamFC} employs cross-correlation to model the relation between exemplar and candidate images. The follow-ups propose upchannel cross-correlation \cite{siamRPN} and depthwise cross-correlation \cite{siamRPNpp} to learn fine-grained feature similarities. Besides their great success, it is important to note that the heuristic matching network design requires substantial effort of human experts, and it is extremely difficult to guarantee robustness in all challenging environments, as experimentally verified in Fig.~\ref{fig:vis} and Tab.~\ref{tab:operators}. One straightforward solution is to find the optimal matching operator under various circumstances, which is however obviously tedious and impractical. Hence, it is natural to throw a question: \emph{can we search for a general matching network for Siamese tracking?}
	 
	 
	 In this work, we show the answer is affirmative by proposing a search algorithm for automatic matching network design. Instead of adopting the conventional cross-correlation and its variants, we explore more feasibility of matching operator selection. Specifically, besides cross-correlation, we introduce six novel matching operators to Siamese tracking, namely Concatenation, Pointwise-Addition, Pairwise-Relation, FiLM, Simple-Transformer and Transductive-Guidance. We shed light on the intrinsic differences of these operators by comparing their performances under different environment degradation types. Surprisingly, by simply replacing the cross-correlation to concatenation, the strong baseline tracker Ocean \cite{Ocean} achieves 1.2 points gains on success score of OTB100 \cite{OTB-2015} (see Tab.~\ref{tab:operators}). Moreover, we observed that the matching operators show different resilience on various challenging factors and image contents. This inspires us to combine them to exploit  complementary informative features.
	 
    To this end, we propose a search algorithm, namely Binary Channel Manipulation (BCM), to automatically select and combine matching operators. Firstly, we construct a search space with the aforementioned seven operators. The exemplar and candidate images pass through all matching operators to generate the corresponding response maps. For each response channel, we assign it with a learnable manipulator to indicate its contribution for other tracking steps. Gumbel-Softmax \cite{gumbel-softmax} is applied to discretize the manipulators as binary decision, as well as guarantee the differentiable training. Then, we aggregate manipulators of all channels to identify the operator's potential for adapting to the baseline tracker. Our search algorithm aims to find the matching networks with better generalization on different tracking environments. Thus, the performance on the validation set is treated as the reward or fitness. Concretely, we solve the search algorithm using bilevel optimization, which finds the optimal manipulators on the validation set with the weight of other layers (\emph{e.g.,} convolution kernels) learned on the training data. Notably, we simultaneously predict matching networks for both the classification and regression branches in state estimation. \textbf{The different search results for classification and regression demonstrate that our method is capable of finding task-dependent matching networks.} Finally, we integrate the learned matching networks into the baseline tracker \cite{Ocean} and train it following the standard Siamese procedure.
    
	 
	 The effectiveness of the proposed framework is verified on OTB100 \cite{OTB-2015}, LaSOT \cite{LASOT}, GOT10K \cite{GOT10K}, TrackingNet \cite{TrackingNet} and TNL2K \cite{TNL2K}. Our approach surpasses the baseline tracker \cite{Ocean} on all five benchmarks. It is worth noting that the proposed tracker also outperforms the recent online updating methods DiMP \cite{DiMP} and KYS \cite{KYS} on all criteria of the evaluated datasets. 
	 
	 The main contributions of this work are twofold.
	 \begin{itemize}[leftmargin=0.55cm]
    \item{
        We introduce six novel matching operators for Siamese tracking. A systematic analysis reveals that the commonly-used (depthwise) cross-correlation is not a requisite, and an appropriate matching operator can further bring remarkable performance gains.
    }

    \item{ 
        A conceptually simple algorithm, namely Binary Channel Manipulation (BCM), is proposed for automatic matching networks design with the introduced operators. By integrating the learned matching networks into the baseline tracker, it achieves remarkable performance gains with neglectable overhead on tracking speed.
    }

\end{itemize}

	\section{Related Work}
	In this section, we review the related work on matching based tracking, as well as briefly describe recent thriving Siamese trackers, where the baseline tracker belongs to.
	
	 
	
	\subsection{Tracking via Heuristic Matching} \label{SEC2-1}
	In the context of visual tracking, it usually corresponds to the process of predicting foreground probability as a one-shot matching problem. SINT \cite{SINT} proposes
	to learn a matching function to identify candidate image locations that match with the initial object appearance. The matching function is simply defined as \emph{dot product} operation. Held et al. introduce GOTURN \cite{GOTURN}, which predicts target location by directly regressing the \emph{concatenation} feature of the exemplar and candidate images. Global Track \cite{GlobalTrack} and ATOM \cite{ATOM} inject the target information into the region proposal network by applying \emph{hadamard product} to exemplar and candidate embeddings. Recent prominent Siamese trackers \cite{siamFC, siamRPN, siamRPNpp} achieve groundbreaking results on all benchmarks, which is mostly attributed to the effective \emph{cross-correlation} module and its variants. We observed that when choosing matching functions for a tracking method, expertise and massive experiments are inevitably required. Moreover, the heuristic matching network may not be an optimal architecture design. In this work, we propose a differentiable search algorithm to automatically determine which matching functions to use and how to combine them in visual tracking. Since the proposed search algorithm is applied to the Siamese framework, in the following, we briefly retrospect the development of Siamese tracking.
	
	

\begin{figure*}[!t]
\vspace{-0.5em}
	\begin{center}
		\includegraphics[width=1\linewidth]{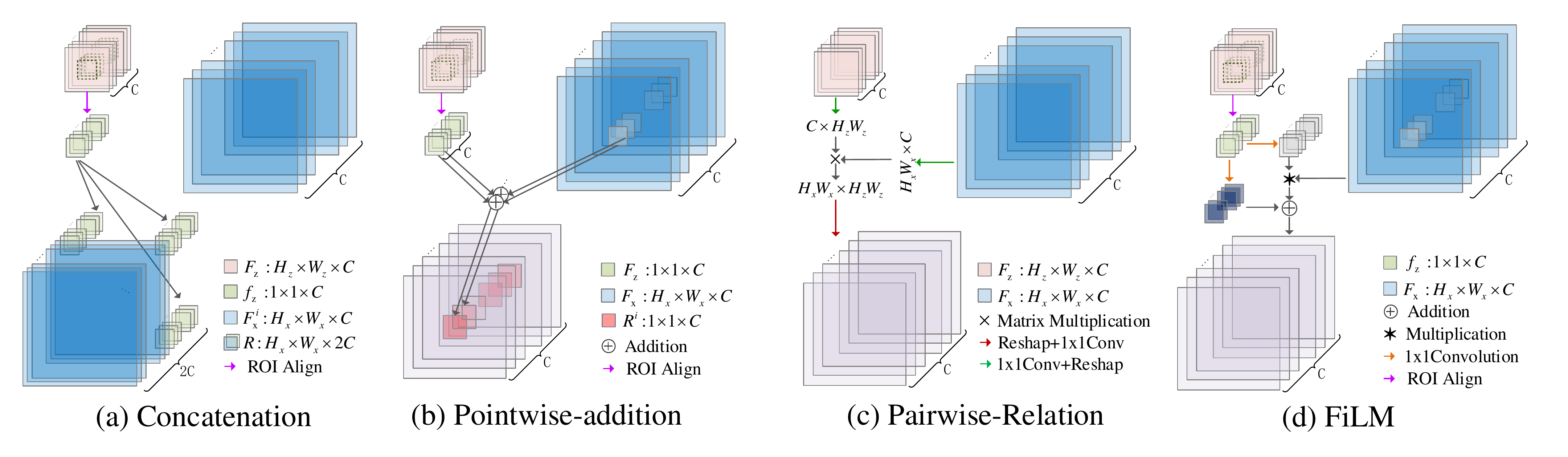}
	\end{center}
    \vspace{-2em}
	\caption{Matching operators: (a) Concatenation (b) Pointwise-Addition (c) Pairwise-Relation (d) FiLM (see Sec.~\ref{sec:Operators}).} 
    \vspace{-0.5em}
	\label{fig:operators1}
\end{figure*}

	\subsection{Siamese Tracking} \label{SEC2-2}
	Siamese tracking has drawn attention because of its balanced accuracy and speed. The pioneering work of Siamese trackers, \emph{i.e.}, SiamFC \cite{siamFC},  introduces the \emph{cross-correlation} layer as a similarity metric for target matching, which significantly boosts tracking efficiency. SiamRPN \cite{siamRPN} ensues to improve SiamFC by advocating a region proposal network for scale estimation. The follow-up works unleash the capability of deeper backbone networks in Siamese tracking by alleviating position bias \cite{siamRPNpp} and perceptual inconsistency \cite{SiamDW}.  The estimation network evolves from anchor-based to anchor-free mechanism recently \cite{SiamBAN,SiamCAR,Ocean,SiamFC++}.  Whilst deeper backbone and advanced estimation network significantly enhance the transferability of tracking models, the feasibility of matching network design remains less investigated. In this work, we narrow this gap by introducing new matching operators and searching their optimal combination for Siamese tracking.

	

\section{Analysis of Matching Operators}
\subsection{Instantiations} \label{sec:Operators}

The standard Siamese tracker takes an exemplar image $\bm{z}$ and a candidate image $\bm{x}$ as input. The image $\bm{z}$ represents object of interest in the first frame, while $\bm{x}$ is typically larger and represents the search area in subsequent video frames. The two images are first fed into a shared backbone network to generate two corresponding feature maps $\bm{F}_{z} \in \mathbb{R}^{H_z \times W_z \times C}$ and $\bm{F}_{x} \in \mathbb{R}^{H_x \times W_x \times C}$. Then a matcing network $\varphi$ is applied to inject the information of exemplar $\bm{F}_{z}$ to $\bm{F}_{x}$, which outputs a correlation feature $\bm{R}$,
\begin{equation}
\bm{R} = \varphi(\bm{F}_{z}, \bm{F}_{x}).
\label{eq:matching}
\end{equation}
Recent top-ranked Siamese trackers define $\varphi$ as \emph{depth-wise cross-correlation} \cite{siamRPNpp, SiamAtt, SiamBAN, SiamCAR, SiamFC++, Ocean}. Notably, when the spatial size of $\bm{F}_z$ is $1\times 1$ ($\bm{f}_z$), the depthwise cross-correlation resembles hadamard product \cite{GlobalTrack}. Besides depthwise cross-correlation, in this work, we explore other matching operators, namely \emph{Concatenation}, \emph{Pointwise-Addition}, \emph{Pairwise-Relation}, \emph{FiLM}, \emph{Simple-Transformer} and \emph{Transductive-Guidance}. The concatenation operator has been exploited in previous work \cite{GOTURN}, while others have not, to the best of our knowledge. We detail each of them in the following.


\textbf{Concatenation} is used by the pairwise function in Relation Networks \cite{RelationNetwork} for visual reasoning. We also explore a concatenation form of $\varphi$, as shown in Fig.~\ref{fig:operators1} (a):
\begin{equation}
\bm{R} = \operatorname{Conv}([\bm{f}_{z}, \bm{F}_{x}]),
\label{eq:concatanation}
\end{equation}
here $\bm{f}_{z} \in \mathbb{R}^{1 \times 1 \times C}$ is the pooled features on $\bm{F}_{z}$ (inside the bounding box).  [·, ·] denotes concatenation and $\operatorname{Conv}$ is a $1\times 1$ convolution layer with output channel of $C$.

\textbf{Pointwise-Addition} is similar to the hadamard product, but changes ``multiplication'' to ``addition'' (see Fig.~\ref{fig:operators1} (b)):
\begin{equation}
\bm{R} = \bm{f}_{z} + \bm{F}_{x},
\label{eq:addition}
\end{equation}
where $+$ denotes elementwise addition.

\begin{table*}[!t]
    \begin{center}

        \caption{Performance (Success Rate) of different operators on OTB100 \cite{OTB-2015}. Illumination Variation (IV), Scale Variation (SV), Occlusion (OCC), Deformation (DEF), Motion Blur (MB), Fast Motion (FM), In-Plane Rotation (IPR), Out-of-Plane Rotation (OPR), Out-of-View (OV), Background Clutters (BC) and Low Resolution (LR) are 11 challenging attributes. } 
        \vspace{0.5em}
        \fontsize{9pt}{4.2mm}\selectfont
        \begin{threeparttable}
            \begin{tabular}{@{}c@{} | @{}l@{} | @{}c@{} | @{}c@{} @{}c@{}  @{}c@{} @{}c@{} @{}c@{} @{}c@{} @{}c@{} @{}c@{} @{}c@{} @{}c@{} @{}c@{} @{}c@{} @{}c@{}}
                \cline{1-14}
                \# NUM ~& ~~\# Operators & ~~Overall ~~& ~~~IV~~ & ~~~SV~~~  & ~~~OCC~~~ & ~~~DEF~~~ & ~~~MB~~~ & ~~~FM~~~ & ~~~IPR~~~ & ~~~OPR~~~ & ~~~OV~~~ & ~~~BC~~~ & ~~~LR~~~
                \\
                \cline{1-14} 
                \ding{172}& ~~Depthwise Cross-correlation ~~& 67.2 & ~69.3~ & ~67.7~ & ~62.8~ &~65.2~&~68.3~& ~67.5~ & ~67.8~ & ~66.6~ & ~\textbf{63.9}~ &~62.6~ &~67.9~
                \\

                \ding{173}& ~~Concatenation & \textbf{68.4} & ~~\textbf{71.5}~~ & ~67.3~ & ~\textbf{65.2}~ & ~\textbf{66.5}~ & ~\textbf{70.0}~ & ~\textbf{69.0}~ & ~\textbf{69.8}~ & ~67.2~ & ~62.7~ & ~\textbf{65.3}~&~65.6~
                \\

                \ding{174}& ~~Pointwise-Addition & 67.1 & ~~66.6~~& ~66.2~& ~61.5~ & ~61.8~ & ~65.6~ & ~66.8~ & ~67.7~ & ~65.9~ & ~52.2~ &~58.1~& ~\textbf{69.7}~
                \\
                \ding{175}& ~~Pairwise-Relation & 67.8 & ~~67.0~~ & ~66.5~ & ~63.7~ & ~65.1~ & ~68.0~ &~66.6~ & ~66.7~ & ~\textbf{68.2}~ & ~57.2~ &~63.6~& ~59.8~ 
                \\
                
                \ding{176}& ~~FiLM & 67.4 & ~~69.4~~ & ~66.9~ & ~60.4~ & ~63.7~ & ~66.9~ &~67.3~ & ~66.8~ & ~65.2~ & ~53.7~ &~58.5~ & ~66.8~
                \\
                
                \ding{177}& ~~Simple-Transformer & 65.8 & ~~67.3~~ & ~65.8~ & ~60.1~ & ~62.1~ & ~65.9~ &~65.7~ & ~66.8~ & ~66.0~ & ~55.4~ & ~60.7~& ~64.8~
                \\
                
               \ding{178} & ~~Transductive-Guidance & 65.0 & ~~64.8~~ & ~\textbf{68.3}~ & ~61.6~ & ~61.2~ & ~67.2~ & ~65.0~ & ~64.9~ & ~65.0~ & ~57.6~ & ~56.0~ & ~64.2~
                \\

                \cline{1-14}

            \end{tabular}

        \end{threeparttable}
        \vspace{-1.5em}

        \label{tab:operators}
    \end{center}
\end{table*}

\textbf{Pairwise-Relation} is widely used in video object segmentation \cite{TVOS}. It is a variant of non-local attention \cite{NonLocal}, and is defined as, 
\begin{equation}
\bm{R} = \operatorname{matmul}(S(\bm{F}_x), S(\bm{F}_z)),
\label{eq:pairwise-relation}
\end{equation}
where $S$ reshapes $\bm{F}_x$ and $\bm{F}_z$ to the size of $H_x W_x \times C$ and $C \times H_z W_z$, respectively (see Fig.~\ref{fig:operators1} (c)). Here, $\operatorname{matmul}$ denotes matrix multiplication. The pairwise-relation measures the affinity of each cell in the exemplar feature to all that in the candidate feature.


\textbf{FiLM} is firstly introduced in visual reasoning \cite{FiLM}. It learns to adaptively influence the output of a neural network by applying an affine transformation to the network’s ``intermediate features'', based on some ``input''. For visual tracking, we consider the exemplar feature $\bm{f}_z$ as the ``input'', and the candidate feature $\bm{F}_x$ as ``intermediate features''. More formally,
\begin{equation} 
\begin{split}
\gamma = \operatorname{Conv}(\bm{f_z}), \\
\beta = \operatorname{Conv}(\bm{f_z}), \\
\bm{R} = \gamma \bm{F}_x + \beta,
\end{split}
\label{eq:FiLM}
\end{equation}
where the coefficient $\gamma$ and bias $\beta$ are two tensors with size of $1 \times 1 \times C$, as shown in Fig.~\ref{fig:operators1} (d). 

\begin{figure}[!t]
	\begin{center}
		\includegraphics[width=1\linewidth]{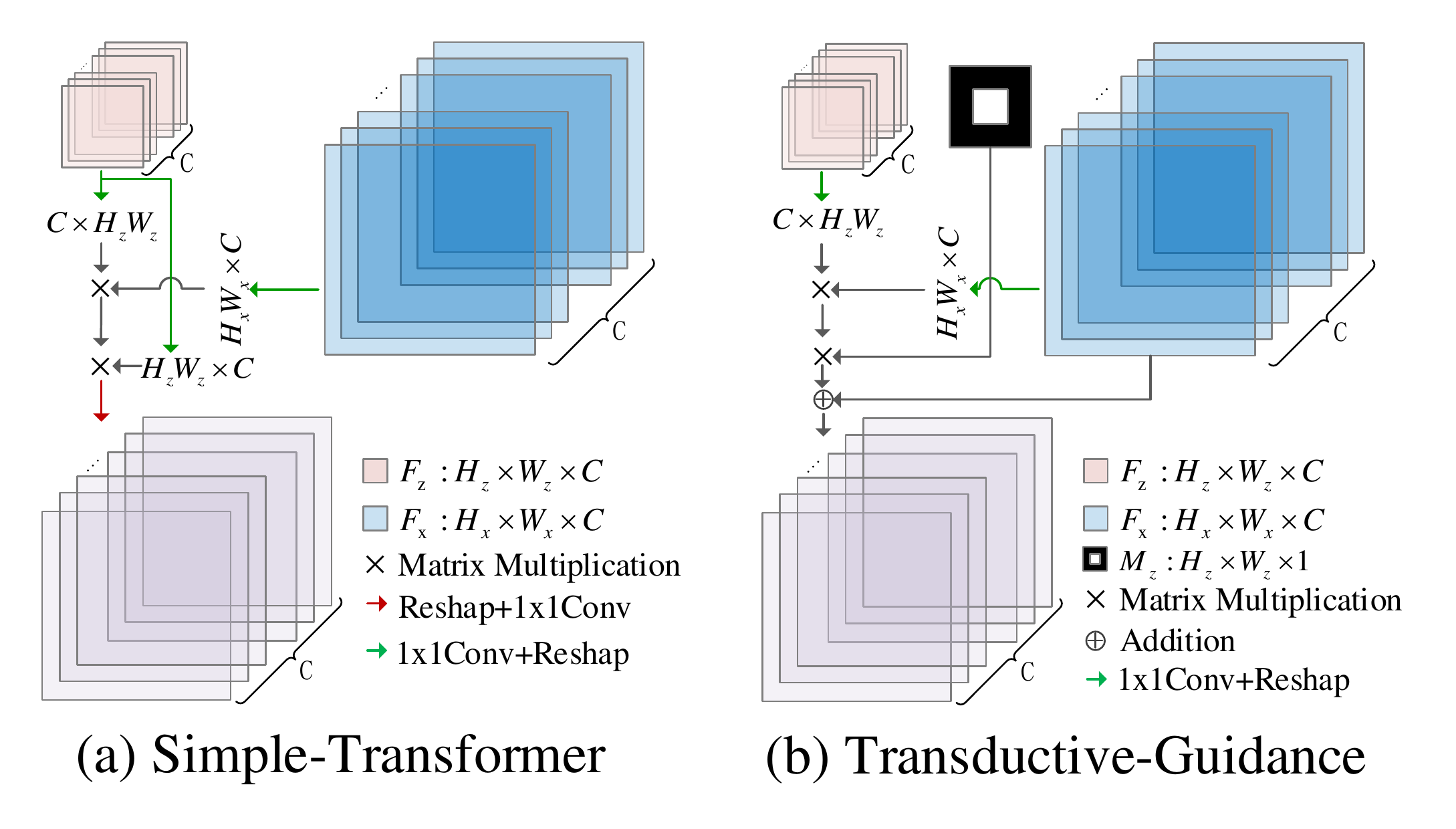}
	\end{center}
    \vspace{-1.5em}
	\caption{Matching operators: (a) Simple-Transformer (b) Transductive-Guidance. Details are described in Sec.~\ref{sec:Operators}.} 
    \vspace{-0.5em}
	\label{fig:operators2}
\end{figure}

\textbf{Simple-Transformer} is motivated by recent booming visual transformer \cite{ViTsurvey}, 
\begin{equation} 
\bm{R} = \operatorname{Att}(query, key, value),
\label{eq:FiLM}
\end{equation}
where $query=\operatorname{Conv}(\bm{F}_x), key=\operatorname{Conv}(\bm{F}_z), value=\operatorname{Conv}(\bm{F}_z)$. $\operatorname{Att}$ is a multi-head attention layer in visual transformer \cite{ViTsurvey}, and is implemented by ``nn.multiheadAttention'' in PyTorch \cite{PYTORCH}. More details are presented in Fig.~\ref{fig:operators2} (a).

\textbf{Transductive-Guidance} is originated from mask propagation mechanism in video object segmentation \cite{TVOS, OceanPlus}, where the segmentation masks of previous frames guide the prediction of the current frame. In our work, we specifically modify it for Siamese tracking. First, the affinity between examplar and candidate feature is predicted by,
\begin{equation}
\bm{A} = \operatorname{matmul}(S(\bm{F}_x), S(\bm{F}_z)). \\	
\end{equation} 
This step is the same as the computation of the pairwise-relation. With the affinity, the spatial guidance is learned by propagating the pseudo mask of the first frame,
\begin{equation}
\bm{G} = \operatorname{matmul}(\bm{A}, S(\bm{M}_z)),
\end{equation} 
where $\bm{M}_z$ is the pseudo mask of the first frame. Specifically, the pixels inside and outside the bounding box are set to 1 and 0, respectively, as shown in Fig.~\ref{fig:operators2} (b). $\bm{G}$ serves as the spatial guidance for target localization, in which each pixel indicates the foreground probability of a location. Then the spatial guidance is fused with the visual feature by,
\begin{equation}
\bm{R} = \bm{G} + \bm{F}_x.
\end{equation}

\begin{figure}[!t]
	\begin{center}
		\includegraphics[width=1\linewidth]{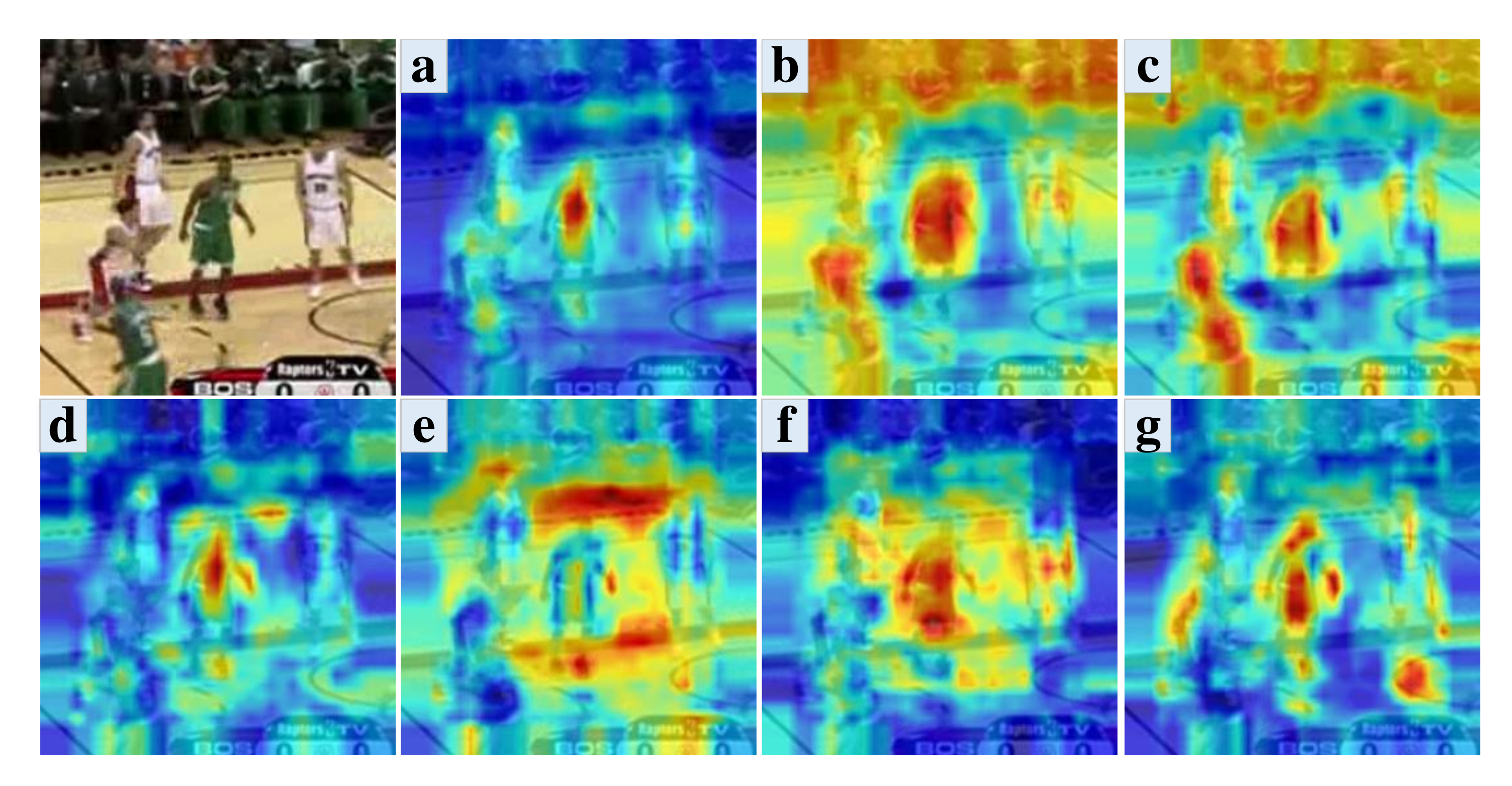}
	\end{center}
    \vspace{-1em}
	\caption{Activation maps of different matching operators. (a) Depthwise Cross-correlation (b) Concatenation (c) Pointwise-Addition (d) Pairwise-Relation (e) FiLM (f) Simple-Transformer (g) Transductive-Guidance.} 
\vspace{-1em}
	\label{fig:quantitative}
\end{figure}

\begin{figure*}[!t]
	\begin{center}
	\hspace{-0.5em}
		\includegraphics[width=1\linewidth]{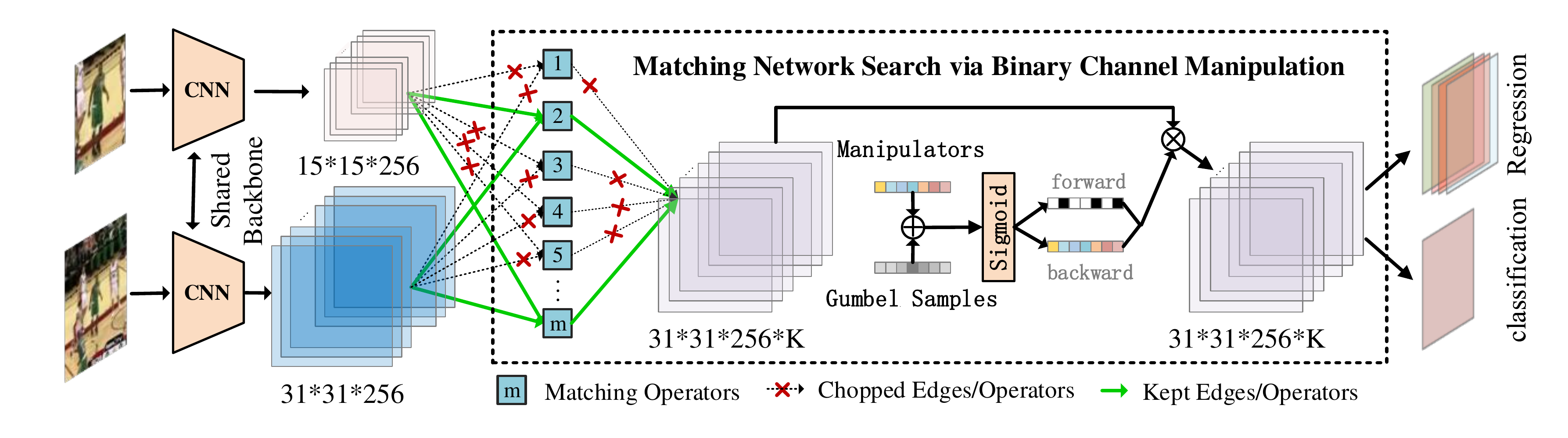}
	\end{center}
    \vspace{-1.5em}
	\caption{Overview of the proposed framework AutoMatch. The \textcolor[RGB]{146,205,220}{matching operators} in search space explore the relation between exemplar and candidate features. The \textcolor[RGB]{192,0,0}{crosses} and dashed arrows indicate the discarded operators after searching with binary channel manipulation. And operators linked with the \textcolor{green}{green arrows} constructs the searched matching network. The search algorithm is applied to both classification and regression, and only one of that is illustrated here for simplicity.}
	
	\label{fig:framework}
\end{figure*}

\subsection{Analysis} \label{sec:CC}

In Sec.~\ref{sec:Operators}, we introduce six novel matching operators for Siamese tracking, besides the conventional depthwise cross-correlation. It is natural to ask: \emph{How do these new operators perform, and could the conventional depthwise cross-correlation be replaced by these proposed operators?} We answer the questions in this section.

\textbf{Performance of Individual Operators.} To investigate the impact of each operator on Siamese tracking, we apply them to a recent tracker Ocean \cite{Ocean}, and evaluate the performance on OTB100 \cite{OTB-2015}. As shown in Tab.~\ref{tab:operators}, the vanilla Ocean \cite{Ocean} with depthwise cross-correlation (\ding{172}) achieves overall success of 67.2. When replacing the depthwise cross-correlation by Simple-Transformer (\ding{177}) and Transductive-Guidance (\ding{178}), the overall score drops to 65.8 and 65.0, respectively. The performance degradation illustrates that randomly choosing a matching operator may bring negative impacts to a tracking framework. But surprisingly, the results of all other four operators (\ding{173}$\sim$ \ding{176}) are favorably comparable to or even better than depthwise cross-correlation. The comparisons inspire us that the classical depthwise cross-correlation is not the optimal choice for Siamese tracking, and an appropriate matching operator can lead to better tracking accuracy. 


\textbf{Potential of Complementarity.} Although one well-designed matching operator may surpass classical depthwise cross-correlation under certain circumstances, the improvements cannot be guaranteed for all challenging cases. As shown in Tab.~\ref{tab:operators}, although the concatenation operator (\ding{173}) exhibits superiority over most challenging factors, it is inferior to Transductive-Guidance (\ding{178}) on Scale Variation (SV), Pairwise-Relation (\ding{175}) on Out-of-Plane Rotation (OPR), Depthwise Cross-correlation (\ding{172}) on Out-of-View (OV) and Point-Addition (\ding{174}) on Low Resolution (LR).  We further visualize the activation map of matching outputs in Fig.~\ref{fig:quantitative}. It shows that the depthwise cross-correlation (a), Pairwise-relation (d), and Transductive-Guidance (g) tend to filter out the context features and focus on the target itself.  Conversely, the concatenation (b), Pointwise-Addition (c), Simple-Transformer (e), and FiLM (e) exploit more context information. The possible reason is that the hard negative examples introduced by the context help prevent overfitting to the easy background.  



In a nutshell, the quantitative comparison in Tab.~\ref{tab:operators} and qualitative analysis in Fig.~\ref{fig:quantitative} demonstrate that different matching operators show different resilience on various challenging factors and image contents. This inspires us to combine them to exploit complementary informative features. Instead of searching for the best matching operators under various circumstances, which is obviously impractical, we propose an automatic method that can adaptively learn to choose and combine the matching functions.

\section{Methodology} \label{sec:method}



\subsection{Overview of AutoMatch} \label{sec:method-overview}
 The proposed framework AutoMatch is illustrated the in Fig.~\ref{fig:framework}. Typical Siamese tracking framework contains three main steps, \emph{i.e.}, feature extraction, matching, and target localization. Given an exemplar image $\bm{z}$ and a candidate image $\bm{x}$, a backbone network is first applied to extract visual features $\bm{F}_z$ and $\bm{F}_x$.  $\bm{F}_z$ and $\bm{F}_x$ then pass through a matching network $\varphi$ to learn their relation. $\varphi$ is generally defined as depthwise cross-correlation in recent works \cite{siamRPNpp, Ocean}. In our study, the matching network design evolves from heuristic selection to automatic search. Concretely, $\bm{F}_z$ and $\bm{F}_x$ are fed to matching operators in the search space (see Sec.~\ref{sec:Operators}), which obtains $m$ multi-channel response features $\{\bm{r}_1, \bm{r}_2, ..., \bm{r}_m\}$. Each channel of a response feature is assigned with a learnable manipulator $w_i^{j}$, indicating a feature channel's contribution to other tracking steps. We introduce the binary Gumbel-Softmax \cite{gumbel-softmax} to discretize the manipulators for binary decision, as well as guarantee the differentiable training. The learning of manipulators is formulated as bilevel optimization (see Sec.~\ref{sec:method-BO}). Two operators are finally retained based on the guidance of the learned manipulators, and their response maps are concatenated as the input of the following steps. With the learned matching networks, the classification and regression networks are followed to predict the target state (see Sec.~\ref{sec:ex-training}). 
 


\subsection{Binary Channel Manipulation} \label{sec:method-BCM}
Let $\mathcal{O}=\{o_1,o_2, ..., o_m\}$ be the search space consisting of optional matching operators $o_i(\cdot)$ to be applied to exemplar and candidate features. The response set $\mathcal{R}$ is got by,
\begin{equation}
    \mathcal{R} = \{o_1(\bm{z}, \bm{x}), ..., o_m(\bm{z}, \bm{x})\}.
    \label{eq:rpset}
\end{equation}
The search algorithm aims to find the optimal combination of operators based on the response set $\mathcal{R}$. We propose binary channel manipulation (BCM) to decide the contribution of an operator for target state prediction. Each element  $\bm{r}_i^j$ in $\mathcal{R}$ is a tensor with size of $H_x \times W_x \times C$. We assign each feature channel with a learnable manipulator $w_i^{j}$, and then aggregates the weighted maps in $\mathcal{R}$ by concatenation,
\begin{equation}
    \bm{E} = [\sigma(w_1^{1})\bm{r}_1^{1},..., \sigma(w_i^{j})\bm{r}_i^{j}, ..., \sigma(w_m^{C})\bm{r}_m^{C}],
    \label{eq:aggre}
\end{equation}
where $\bm{r}_i^{j}$ indicates the $j_{th}$ channel of the $i_{th}$ response feature. $\sigma$ is sigmoid. $\bm{E} \in \mathbb{R}^{H_x \times W_x \times C|\mathcal{O}|}$ denotes the aggregated feature, which is used as the input of subsequent target estimation network. The manipulator defines the channel's contribution to target location. For each operator, we define the summation of channel manipulators as the potential $p_i$ of an operator for adapting to the baseline tracker,
\begin{equation}
    p_i = \sum_{j=1}^{C} \sigma(w_i^{j}).
    \label{eq:adaption}
\end{equation}

Inspired by channel pruning \cite{GATEDNETWORKS} and differentiable network architecture search \cite{FairDARTS, DARTS}, we translate the continuous solution $w_i^{j}$ to discrete one for final decision. These discrete decisions are trained end-to-end using the Gumbel-Softmax \cite{gumbel-softmax}. Concretely, given a distribution with (two) class probabilities $\pi=\{\pi_1=\sigma(w_i^{j}), \pi_2=1-\sigma(w_i^{j})\}$, the discrete samples $d$ can be drawn using,
\begin{equation}
    d = \operatorname{onehot}(\mathop{\arg\min}_{k}[\log(\pi _{k}) + g_k]),
\end{equation}
where $g_k$ is noise sample drawn from Gumbel distribution. $k \in \{1,2\}$ denotes binary classification. The Gumbel-Softmax defines a continuous, differentiable approximation by replacing the argmax with a softmax,
\begin{equation}
    y_k = \frac{\exp((\log(\pi _k)+g_k) / \tau)}{\sum_{c=1}^{2}\exp((\log(\pi _c)+g_c) / \tau)}.
    \label{eq:GumbelCon}
\end{equation}
Substituting $\pi _{1}=\sigma(w_i^{j})$, $\pi _{2}=1-\sigma(w_i^{j})$, Eq. \ref{eq:GumbelCon} is simplified to ($k=1$ for binary case),
\begin{equation}
    y_1 = \sigma(\frac{w_i^{j}+g_1-g_2}{\tau}).
    \label{eq:GumbelFinal}
\end{equation}
We attach the derivation in supplementary materials due to space limit. The $\tau$ is set to 1, $g_k$ to 0 following \cite{BengioConditional, gumbel-softmax}. For the  discrete sample $d$, a hard value is used during the forward pass and gradients are obtained from soft value during the backward pass:
\begin{equation}
d=\left\{
             \begin{array}{lr}
             y_1 > 0.5 \equiv \frac{w_i^{j}+g_1-g_2}{\tau} = w_i^{j} >0,  forward&  \\
             y_1, backward.&  
             \end{array}
\right.
\label{GumbelDis}
\end{equation}

\begin{table*}[!t]
	
	\begin{center}
		\fontsize{9.5}{9.5}\selectfont
		\caption{Result comparisons on five tracking benchmarks. The \textcolor{red}{red}, \textcolor[RGB]{0,139,69}{green} and \textcolor{blue}{blue} indicate performances ranked at the first, second, and third places. Ocean \cite{Ocean} is our baseline model, and we apply the proposed search algorithm on it.
		}
		\begin{tabular}{llccccccccccc}
			
			\toprule
			\multirow{2}{*}{Methods}&\multirow{2}{*}{Year}&
			\multicolumn{2}{c}{OTB100}&\multicolumn{2}{c}{LaSOT}&\multicolumn{2}{c}{TrackingNet}&\multicolumn{2}{c}{TNL2K}&\multicolumn{3}{c}{GOT10K}\cr
			\cmidrule(lr){3-4} \cmidrule(lr){5-6} \cmidrule(lr){7-8} \cmidrule(lr){9-10} \cmidrule(lr){11-13}
			& &Succ.&Prec.&Succ.&Prec.&Succ. &Prec.&Succ.&Prec.&AO&SR$_{0.5}$&SR$_{0.75}$\cr
			\midrule
			SiamFC~\cite{siamFC}&2016&58.7&77.2&33.6&33.9&57.1&66.3& 29.5&28.6 &34.8&35.3&9.8 \cr
			MDNet~\cite{MDNet}&2016&67.8&90.9&39.7&37.3&60.6&56.5&31.0&32.2&29.9&30.3&9.9  \cr
			ECO~\cite{ECO}&2017&69.1&91.0&32.4&30.1&55.4&49.2&32.6&31.7&31.6&30.9&11.1 \cr
			VITAL~\cite{VITAL}&2018&69.1&91.7&39.0&36.0&-&-&36.6&35.3&35.0&36.0&9.0 \cr
			GradNet~\cite{GradNet}&2019&63.9&86.1&36.5&35.1&-&-&31.7&31.8&-&-&- \cr
			SiamDW~\cite{SiamDW}&2019&67.4&90.5&38.4&35.6&-&-&32.3&32.6&41.6&47.5&14.4\cr
			SiamRPN++~\cite{siamRPNpp}&2019&69.6&92.3&49.6&49.1&73.3&69.4&41.3&41.2&51.7&61.6&32.5\cr
			ATOM~\cite{ATOM}&2019&66.7&87.9&51.5&50.5&70.3&64.8&40.1&39.2&55.6&63.4&40.2 \cr
			DiMP~\cite{DiMP}&2019&68.6&89.9&\textcolor[RGB]{0,139,69}{56.9}&\textcolor[RGB]{0,139,69}{56.7}&74.0&68.7&\textcolor{blue}{44.7}&\textcolor[RGB]{0,139,69}{43.4}&\textcolor{blue}{61.1}&\textcolor{blue}{71.7}&\textcolor{blue}{49.2} \cr
			SiamFC++~\cite{SiamFC++}&2020&68.3&91.2&54.3&54.7&\textcolor{blue}{75.4}&\textcolor{blue}{70.5}&38.6&36.9&59.5&69.5&47.9 \cr
			D3S~\cite{D3S}&2020&-&-&-&-&72.8&66.4&38.8&39.3&59.7&67.6&46.2 \cr
			MAMLTrack~\cite{MAMLTrack}&2020&\textcolor[RGB]{0,139,69}{71.2}&\textbf{\textcolor{red}{92.6}}&52.3&53.1&\textcolor[RGB]{0,139,69}{75.7}&\textcolor[RGB]{0,139,69}{72.5}&28.4&29.5&-&-&-\cr
			SiamAttn~\cite{SiamAtt}&2020&\textcolor[RGB]{0,139,69}{71.2}&\textbf{\textcolor{red}{92.6}}&\textcolor{blue}{56.0}&-&75.2&-&-&-&-&-&-\cr
			SiamCAR~\cite{SiamCAR}&2020&-&-&50.7&51.0&-&-&35.3&38.4&56.9&67.0&41.5 \cr
			SiamBAN~\cite{SiamBAN}&2020&69.6&91.0&51.4&52.1&-&-&41.0&41.7&-&-&-\cr
			KYS~\cite{KYS}&2020&69.5&91.0&55.4&\textcolor{blue}{55.8}&74.0&68.8&\textcolor[RGB]{0,139,69}{44.9}&\textbf{\textcolor{red}{43.5}}&\textcolor[RGB]{0,139,69}{63.6}&\textcolor[RGB]{0,139,69}{75.1}&\textcolor[RGB]{0,139,69}{51.5}\cr
			Ocean~\cite{Ocean}&2020&67.2&90.2&52.6&52.6&70.3&68.8&38.4&37.7&59.2&69.5&46.5 \cr
			\cmidrule(lr){1-13}
			\textbf{AutoMatch}&Ours&\textbf{\textcolor{red}{71.4}}&\textbf{\textcolor{red}{92.6}}&\textbf{\textcolor{red}{58.3}}&\textbf{\textcolor{red}{59.9}}&\textbf{\textcolor{red}{76.0}}&\textbf{\textcolor{red}{72.6}}&\textbf{\textcolor{red}{47.2}}&\textbf{\textcolor{red}{43.5}}&\textbf{\textcolor{red}{65.2}}&\textbf{\textcolor{red}{76.6}}&\textbf{\textcolor{red}{54.3}}\cr
			\bottomrule
		\end{tabular}
		\vspace{-1em}
		\label{tab:results}
	\end{center}
\end{table*}

\subsection{Bilevel Optimization} \label{sec:method-BO}
With binary channel manipulation, our goal is to jointly learn the manipulators $w$ and the weights $\theta$ of other layers (\emph{e.g.,} convolution layers in operators). Analogous to differentiable architecture search \cite{DARTS}, where the validation set performance is treated as the reward or fitness, we aim to optimize the validation loss. Let $\mathcal{L}_{train}$ and $\mathcal{L}_{val}$ denote the training and validation loss, respectively. The goal for matching network search is to find $w^{*}$ that minimizes the validation loss $\mathcal{L}_{val}(\theta^{*}; w^{*})$, where the network parameters $\theta^{*}$ associated with the architecture are obtained by minimizing the training loss $\theta^{*} = \operatorname{argmin}_w \ \mathcal{L}_{train}(\theta, w^{*})$. This implies a bilevel optimization problem \cite{DARTS, FairDARTS} with
$w$ as the upper-level variable and $\theta$ as the lower-level variable,

\begin{gather}
\operatorname{min}_{w} \  \mathcal{L}_{val}(\theta^{*}(w); w), \\
s.t. \quad \theta^{*}(w)=\operatorname{argmin}_{\theta} \ \mathcal{L}_{train}(\theta, w).
\end{gather}
To speed up the bilevel optimization during training, Liu et.al propose a simple approximation in \cite{DARTS},
\begin{gather}
    \nabla_w  \mathcal{L}_{val}(\theta^{*}(w); w) \\
    \approx \nabla_w  \mathcal{L}_{val}(\theta - \epsilon \nabla_\theta  \mathcal{L}_{train}(\theta, w), w),
    \label{eq:biapp}
\end{gather}
where $\epsilon$ is the learning rate for a step of inner optimization. The derivation is beyond the scope of this work. We refer the reader to \cite{DARTS} for more details about the approximation.

In summary, we propose binary channel manipulation to identify the contribution of a matching operator. Then we learn the manipulators by bilevel optimization. We simultaneously apply the search algorithm on the classification and regression branches in state estimation to learn task-dependent matching networks. After training, the first two operators with the maximum potential $p_i$ are retained (see \textcolor{green}{green arrows} in Fig.~\ref{fig:framework}). Finally, we follow the procedure of the baseline tracker \cite{Ocean} to train the searched architecture.

\section{Experiments} 

\subsection{Implementation Details}\label{sec:ex-training}
\noindent \textbf{Network Architecture.} We adopt the recent Siamese tracker Ocean \cite{Ocean} as the baseline model. The backbone network is the modified ResNet50 \cite{MCF}. The target localization network consists of a classification branch and a regression branch. Though the updating branch of Ocean \cite{Ocean} is not used in our work, our tracker remarkably outperforms its online updating version. We refer the readers to \cite{Ocean} for more details about the baseline tracker. In this work, we simultaneously search for the target-dependent matching networks for the classification and regression branches.

\noindent \textbf{Training Procedure.} The training procedure consists of two stages, \emph{i.e.,} matching network search and new tracker training. In the first stage, we search for the matching networks using methods in Sec.~\ref{sec:method} and determine the best cell based on the validation performance. In the second stage, we use the optimized matching networks to construct a new tracker on the baseline approach Ocean \cite{Ocean}. Both stages are trained with Youtube-BB \cite{YTB}, ImageNet-VID \cite{VID}, ImageNet-DET \cite{VID}, GOT10K \cite{GOT10K} and COCO \cite{COCO} (including training and validation sets). The search algorithm's training takes 5 epochs, with each containing  $6 \times 10^{5}$ pairs. The learning rate exponentially decays from $10^{-3}$ to $10^{-4}$. The training of the new tracker follows the baseline model \cite{Ocean}. \textbf{Notably, we simplify Ocean \cite{Ocean} by reducing the training epochs from 50 to 20 to expedite the learning process.}  For the first 5 epochs, we start with a warmup learning rate of $10^{-3}$. For the remaining epochs, the learning rate exponentially decays from $5 \times 10^{-3}$ to $5 \times 10^{-5}$. Both stages are trained with synchronized SGD \cite{SGD} on 4 GTX2080 Ti GPUs, with each hosting 32 images. 


\subsection{State-of-the-art Comparison}
The search algorithm determines different matching networks for the classification and regression branches. \textbf{After the first stage training, Simple-Transformer and FiLM are retrained for the classification branch, meanwhile, FiLM and Pairwise-Relation are preserved for the regression branch.} We compare the new tracker with state-of-the-art models on five benchmarks. Our tracker achieves compelling performance while running at over 50 FPS. Notably, it only takes less than 24 hours for the second stage training (with 4 GTX2080Ti GPUs), which provides a strong but efficient baseline for further research.


\noindent \textbf{OTB100 \cite{OTB-2015}.} OTB100 is a classical tracking benchmark consisting of 100 sequences. Methods are ranked by the area under the success curve (AUC) and precision (Prec.). As shown in Tab.~\ref{tab:results}, our model achieves the top-ranked AUC score, which outperforms the previous best result by SiamAttn \cite{SiamAtt}, \emph{i.e.}, 71.4 \emph{vs} 71.2. When equipping the baseline tracker Ocean \cite{Ocean} with our searched matching network, it brings favorabale 4.2 points gains, \emph{i.e.}, 71.4 \emph{vs} 67.2. The proposed model also surpasses online updating models ATOM \cite{ATOM}/DiMP\cite{DiMP} for 4.5/2.6 points, respectively.



\noindent \textbf{LaSOT \cite{LASOT}.}
LaSOT is a tracking benchmark designed for long-term tracking. Tab.~\ref{tab:results} shows the comparison results on 280 testing videos. Our method achieves the best AUC and precision score, outperforming Ocean \cite{Ocean} for 5.7 and 7.3 points, respectively. Compared with DiMP \cite{DiMP}, our method achieves improvements of 1.4 points on success score. Notably, the proposed tracker runs at 50 FPS, which is comparable to 58 FPS of Ocean, and faster than 43 FPS of DiMP. The comparisons demonstrate that the proposed method brings significant performance gains with small overhead.
\begin{figure}
\begin{center}
    \vspace{-1em}
	\includegraphics[width=1\linewidth]{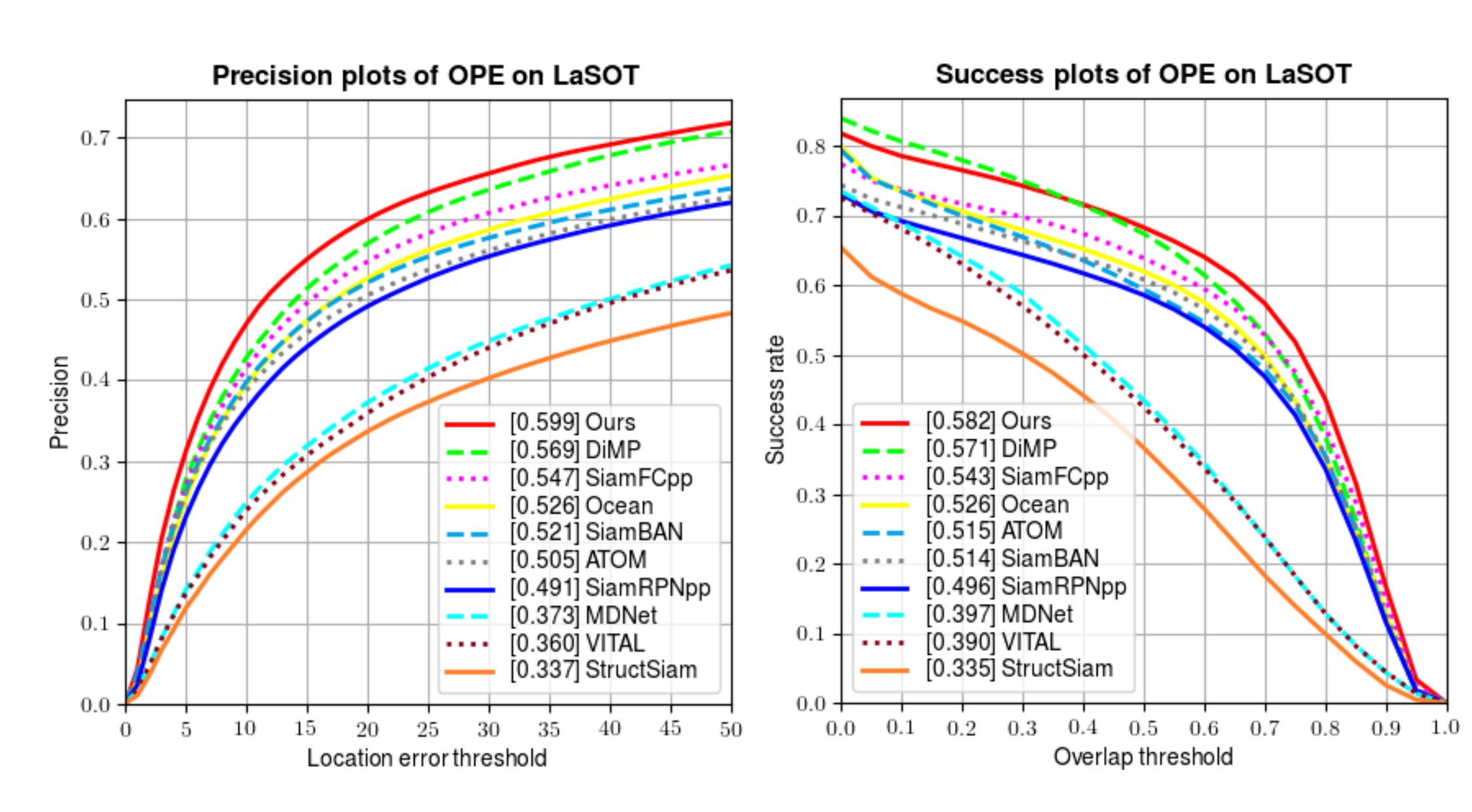}
\end{center}
\vspace{-2em}
\caption{Visualization of results comparison on LaSOT.} 
\vspace{-1em}
\label{fig:lasot}
\end{figure}

\noindent \textbf{TrackingNet \cite{TrackingNet}.}
TrackingNet is a large-scale tracking dataset consisting of 511 sequences for testing. The evaluation is performed on the online server. We report the results in Tab.~\ref{tab:results}. Compared with the baseline tracker Ocean \cite{Ocean}, it achieves 5.7 points gains on success score. Our model also surpasses the meta-learning based MAMLTrack \cite{MAMLTrack} on TrackingNet, \emph{i.e.,} success score of 76.0 \emph{vs} 75.7.

\noindent \textbf{GOT10K \cite{GOT10K}.} 
The evaluation of GOT10K is on the online server. We report the average overlap (AO), success rate (SR$_{0.5}$, SR$_{0.75}$) in Tab.~\ref{tab:results}. Comparing the proposed model with the baseline Ocean \cite{Ocean}, we achieve gains of 6 points, 7.1 points, and 7.8 points on AO, SR$_{0.5}$, and SR$_{0.75}$, respectively. Notably, our model outperforms SiamBAN \cite{SiamBAN} for 1.6 points on AO, while running faster (50FPS \emph{vs.} 40FPS).


\noindent \textbf{TNL2K \cite{TNL2K}.} 
TNL2K is a new dataset which consists of 2000 high diversity videos for natural language guided tracking. Adversarial samples and thermal images are introduced to improve the generality of tracking evaluation. Besides tracking by natural language, it also provides the results of tracking by bounding boxes. In Tab.~\ref{tab:results}, we present the results on 700 testing sequences. It shows that our model achieves the best success and precision scores among the compared trackers.




\subsection{Ablation and Analysis}

\noindent \textbf{One or Many Manipulators.} We link each channel in an operator with a manipulator. Differently, in differentiable neural network search \cite{DARTS}, an operator is identified by a scalar. We also try this strategy, \emph{i.e.,} assigning a matching operator with a scalar during the search. We achieves a final success score of 69.5 on OTB100 \cite{OTB-2015} and 54.7 on LaSOT \cite{LASOT}. The results are inferior to our model, which demonstrates the superiority of our search algorithm. We conjecture that the aggregation of channel information can provide finer guidance for operator selecting.


\noindent \textbf{Random Search.} To demonstrate the efficacy of the search algorithm, we evaluate the performance of random search. Two operators are randomly retained for classification and regression branches, respectively. We report the average performance of the three-time random search and training. The average success score on OTB100 and LaSOT are 69.1 and 53.2. The results manifest that the introduced search method is effective in finding better operators combination.

\begin{figure}[t]
    \centering    
    \vspace{-0.5em}
    \subfloat
    {
        \begin{minipage}[t]{1\textwidth}
            \includegraphics[width=0.48\textwidth]{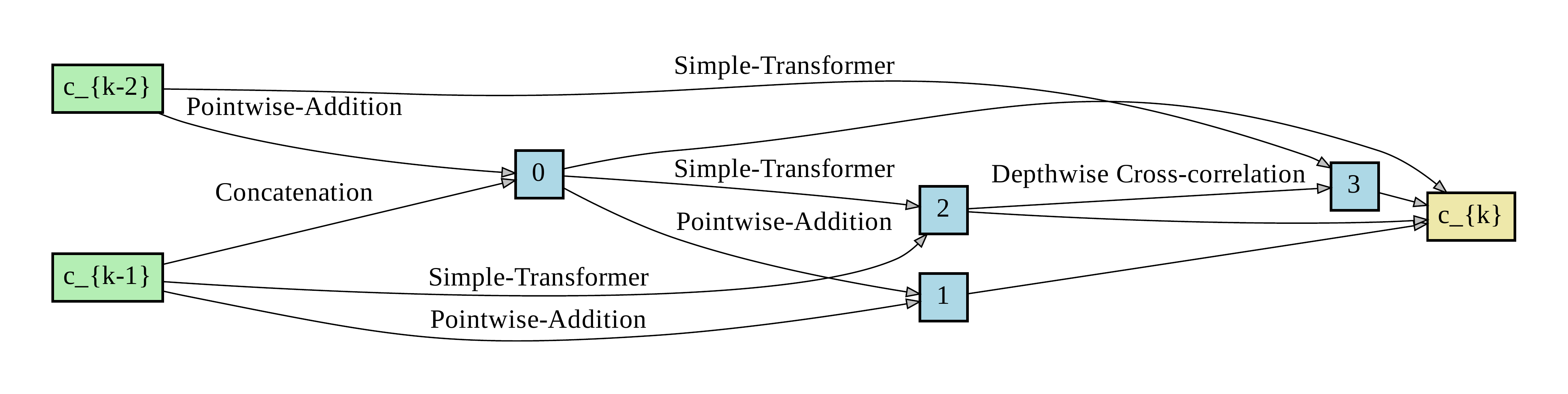}  
        \end{minipage}%
    }
    \vspace{-1.5em}
    \subfloat
    {
        \begin{minipage}[t]{1\textwidth}
            \includegraphics[width=0.48\textwidth]{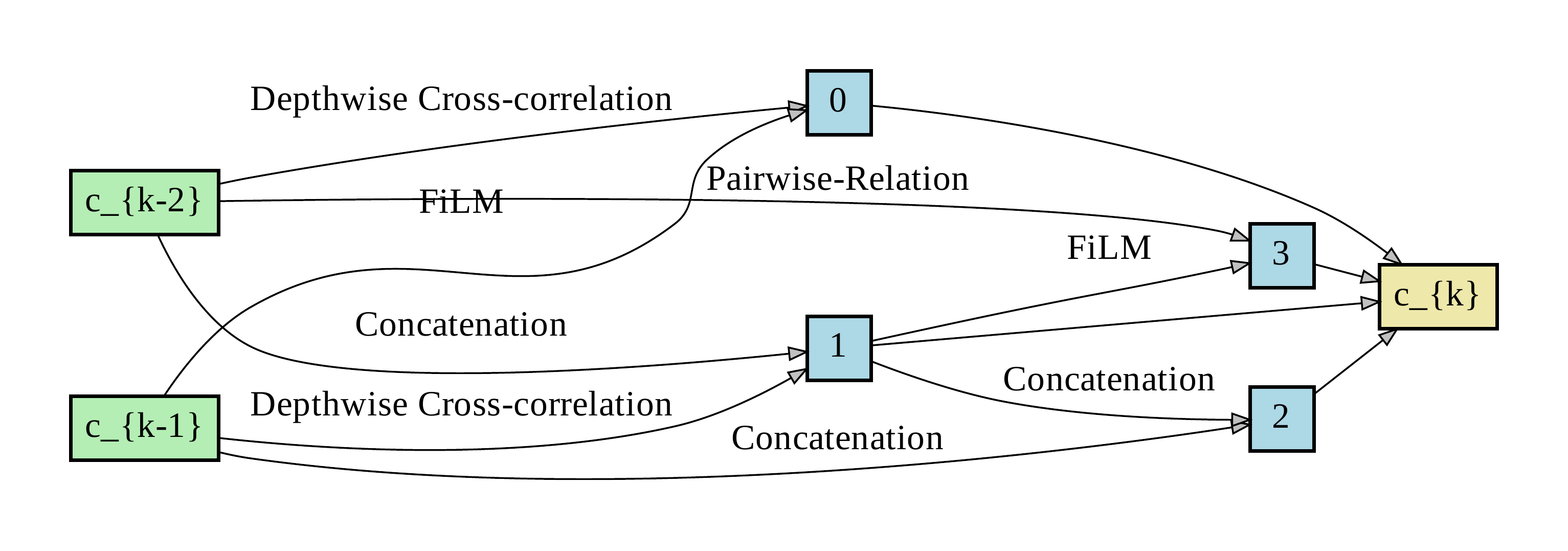}  
        \end{minipage}
    }%
    
    \vspace{-0.5em}
    \caption{\textbf{top:} NAS-like Matching Network for classification. \textbf{bottom:} NAS-like Matching Network for regression.} 
    \label{fig:nas} 
    \vspace{-1.5em}
\end{figure}

\noindent \textbf{NAS-like Matching Cell.} In differentiable neural network search \cite{DARTS}, it represents the basic operating cell as a Directed Acyclic Graph (DAG). Each cell contains multiple nodes, and each node aggregates the outputs of multiple basic operators (\emph{e.g.,} $3\times 3$ convolution layer). One intuitive idea is directly replacing the operators in NAS with our designed matching functions and then searching a matching network. As shown in Fig.~\ref{fig:nas}, we use DARTS \cite{DARTS} to search a matching cell, which looks like that in NAS. Surprisingly, though the searched cell is much complex than ours, it does not show superiority. Concretely, the NAS-like cell achieves an success  score of 55.7 on LaSOT and runs at 35 FPS. Both the performance and inference speed is inferior to the proposed model. The comparison proclaims that directly borrowing NAS to matching network search may not be an optimal choice. We present more details about the DARTS-like structure search and the related work in supplementary materials, due to space limit.

\section{Conclusion}
In this work, we introduce six novel operations to explore more feasibility on matching operator selection in Siamese tracking. Quantitative and quantitative analyses demonstrate that the classical (depthwise) cross-correlation is not the optimal choice for Siamese tracking. We simultaneously find the optimal matching networks for both classification and regression branches in state estimation with the proposed binary channel manipulation (BCM). The learned matching networks are applied to a baseline tracker, and the experimental result shows the robustness of our approach on both short-term and long-term benchmarks. In the future work, we will apply our method to other matching based frameworks, \emph{e.g.,} ATOM.

\noindent \textbf{Acknowledgements.} We thank Heng Fan for his help during ICCV2021 rebuttal. This work was supported by the National Key Research and Development Program of China (Grant No. 2020AAA0106800), the Natural Science Foundation of China (Grant No. 61902401, No. 61972071, No. 61906052, No. 62036011,  No. 61721004, No. 61972394, and No. U2033210), the CAS Key Research Program of Frontier Sciences (Grant No. QYZDJ-SSWJSC040), the Postdoctoral Innovative Talent Support Program BX20200174, China Postdoctoral Science Foundation Funded Project 2020M682828. The work of Bing Li was also supported by the Youth Innovation Promotion Association, CAS.


	
	{\small
		\bibliographystyle{ieee_fullname}
		\bibliography{egbib}
	}
	
\end{document}


	\title{Learn to Match: Automatic Matching Network Design for Visual Tracking  \\
------ Supplementary Material ------}
	
	
	\ificcvfinal\thispagestyle{empty}\fi

    ~\\
    
    \centerline{\textbf{\Large{Learn to Match: Automatic Matching Network Design for Visual Tracking}}}
    \vspace{0.2em}
    \centerline{\textbf{\Large{------ Supplementary Material ------}}}
    
    \vspace{2em}
    
    \centerline{\large{Paper ID 6078}}
    \vspace{2em}

	The supplementary material presents additional details of Sec.4 and Sec.5 in the main manuscript.
	
	$\bullet$ \quad \textbf{(Sec.~4) Derivation of the Binary Channel Manipulation.}
	
	\quad \quad We detail the derivation of Eq.14 $\rightarrow$ Eq.15 in the manuscript.
	
	$\bullet$ \quad \textbf{(Sec.~5) Search with DARTS \cite{DARTS}.}
	
	\quad \quad We provide the modification of searching a DARTS-like matching cell.
	
	$\bullet$ \quad \textbf{Video Demo.}

	\quad \quad We provide a video demo for results comparison of SiamRPN++ \cite{SiamRPNpp}, Ocean \cite{Ocean} and our tracker.

\vspace{1em}
\noindent	\textbf{(Sec.~4) Derivation of the Binary Channel Manipulation.}

In our paper, we simplify the derivation from Eq.14 to Eq.15 due to the space limit. Here we present the details. We rewrite the Eq.14,

\begin{equation}
    y_k = \frac{\exp((\log(\pi _k)+g_k) / \tau)}{\sum_{c=1}^{2}\exp((\log(\pi _c)+g_c) / \tau)}.
    \label{eq:GumbelCon}
\end{equation}
$k=1$ for binary case. Substituting $\pi _{1}=\sigma(w_i^{j})$, $\pi _{2}=1-\sigma(w_i^{j})$, we get,

\begin{align}
    y_1 &= \frac{\exp((\log \sigma(w_i^{j})+g_1) / \tau)}{\exp((\log \sigma(w_i^{j})+g_1) / \tau) + \exp((\log(1-\sigma(w_i^{j}))+g_2) / \tau)} \\
    &= \frac{(\exp(\log \sigma(w_i^{j})+g_1))^{\frac{1}{\tau}}}{(\exp(\log \sigma(w_i^{j})+g_1))^{\frac{1}{\tau}} + (\exp(\log(1-\sigma(w_i^{j}))+g_2))^{\frac{1}{\tau}}}
    \\
    &= \frac{(\exp(\log \sigma(w_i^{j}))\exp(g_1))^{\frac{1}{\tau}}}{(\exp(\log \sigma(w_i^{j}))\exp(g_1))^{\frac{1}{\tau}} + (\exp(\log(1-\sigma(w_i^{j}))\exp(g_2))^{\frac{1}{\tau}}}
    \\
    &= \frac{(\sigma(w_i^{j})\exp(g_1))^{\frac{1}{\tau}}}{(\sigma(w_i^{j})\exp(g_1))^{\frac{1}{\tau}} + ((1-\sigma(w_i^{j}))\exp(g_2))^{\frac{1}{\tau}}}
    \\
    &= \frac{1}{1 + \frac{((1-\sigma(w_i^{j}))\exp(g_2))^{\frac{1}{\tau}}}{(\sigma(w_i^{j})\exp(g_1))^{\frac{1}{\tau}}}}
    \\
    &= \frac{1}{1 + (\frac{1-\sigma(w_i^{j})}{\sigma(w_i^{j})}\exp(g_2 - g_1))^{\frac{1}{\tau}}}
    \\
    &= \frac{1}{1 + (\frac{1-\frac{1}{1+\exp(-w_i^{j})}}{\frac{1}{1+\exp(-w_i^{j})}}\exp(g_2 - g_1))^{\frac{1}{\tau}}}
    \\
    &= \frac{1}{1 + (\exp(-w_i^{j})\exp(g_2 - g_1))^{\frac{1}{\tau}}}
    \\
    &= \frac{1}{1 + \exp(\frac{-w_i^{i}+g_2-g_1}{\tau})}
    \\
    &= \sigma(\frac{w_i^{i}+g_1-g_2}{\tau})
    \label{eq:der} 
\end{align}

\begin{figure}[h]
\begin{center}
    \vspace{-1em}
	\includegraphics[width=0.5\linewidth]{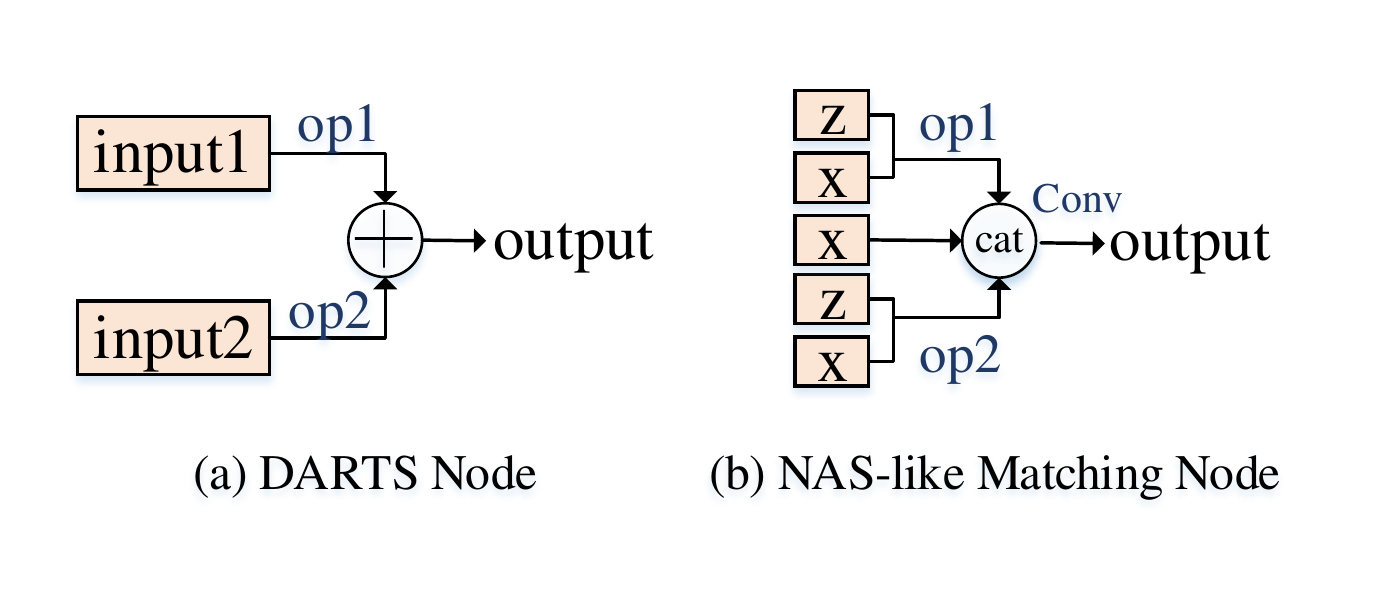}
\end{center}
\vspace{-2em}
\caption{Modify DARTS node for Siamese tracking.} 
\vspace{-1em}
\label{fig:nasdiff}
\end{figure}

\noindent	\textbf{(Sec.~5) Search with DARTS \cite{DARTS}.}
In the section of ``NAS-like Matching Cell'', we present the ablation results of searching with DARTS. DARTS \cite{DARTS} defines the basic cell as a Directed Acyclic Graph (DAG) (see Fig.7 in the main paper). In the original DARTS, each node (the \textcolor[RGB]{0,197,205}{cyan} block in Fig.7 of main paper) links with two output features from previous nodes. After processed by basic operators (\emph{e.g.,} $3\times 3$ convolution), the two inputs of a node are fused by addition (see Fig.~\ref{fig:nasdiff}(a)). However, this mechanism is not suitable for our matching network: 1) The output features of different matching operators have fundamentally different physical meanings. For instance, \emph{Pairwise-Relation} models the global similarity of exemplar and candidate features. Each element in its output represents an affinity score. Nevertheless, the output of \emph{Addition} is the combination of two visual features. The visual feature and similarity feature lie in the different embedding domains. It is thus unreasonable to direct fuse them by addition. In our work, we replace addition with concatenation, and then a $1\times 1$ convolution layer is followed to explore complement of the fused feature.  2) The input of basic operator in DARTS requires only one tensor, yet, we need to feed two tensors into the proposed matching operators, \emph{i.e.,} exemplar and candidate embeddings. The modification is illustrated in Fig.~\ref{fig:nasdiff}. It is worthy to note that we retain the original candidate feature during concatenation. The underlying reason is that involving the visual feature can prevent over-reliance on the relation feature when predicting target locations. Once the relation feature brings ambiguous prediction, \emph{e.g.,} false-positives caused by distractors, the original visual feature with additional information may help to rectify it.

	{\small
		\bibliographystyle{ieee_fullname}
		\bibliography{egbib}
	}